\documentclass[11pt,a4paper]{article}
\usepackage[hyperref]{emnlp-ijcnlp-2019}
\usepackage{times}
\usepackage{latexsym}

\usepackage{url}

\usepackage{amsmath}
\usepackage{multirow}
\usepackage{url}
\usepackage{amsfonts}
\usepackage{enumerate}
\usepackage{graphicx}
\usepackage{array}
\usepackage[ruled]{algorithm2e}
\usepackage{CJKutf8}
\graphicspath{ {images/} }

\aclfinalcopy % Uncomment this line for the final submission

\DeclareMathOperator*{\softmax}{softmax}

\DeclareMathOperator{\LSTM}{LSTM}

\title{Emotion Detection with Neural Personal Discrimination}

\author{Xiabing Zhou${^1}$, Zhongqing Wang${^1}$\thanks{\quad corresponding auther, this paper has been accepted by EMNLP 2019}, Shoushan Li${^1}$, \\
	\textbf{Guodong Zhou${^1}$, Min Zhang${^1}$} \\
	$^1$School of Computer Scienceand Technology, Soochow University, China \\
	{\tt wangzq.antony@gmail.com,} \\
	{\tt \{zhouxiabing, lishoushan, gdzhou, minzhang\}@suda.edu.cn} \\}

\date{}

\begin{document}
\begin{CJK}{UTF8}{gbsn}
\maketitle
\begin{abstract}
There have been a recent line of works to automatically predict the emotions of posts in social media.
Existing approaches consider the posts individually and predict their emotions independently.
Different from previous researches, we explore the dependence among relevant posts via the authors' backgrounds, since the authors with similar backgrounds, e.g., \emph{gender}, \emph{location}, tend to express similar emotions.
However, such personal attributes are not easy to obtain in most social media websites, and it is hard to capture attributes-aware words to connect similar people.
Accordingly, we propose a Neural Personal Discrimination (NPD) approach to address  above challenges by determining  personal attributes from posts, and connecting  relevant posts with similar attributes to jointly learn their emotions.
In particular, we employ adversarial discriminators to determine the personal attributes, with attention mechanisms to aggregate attributes-aware words. In this way, social correlationship among different posts can be better addressed.
Experimental results show the usefulness of personal attributes, and the effectiveness of our proposed NPD approach in capturing such personal attributes with significant gains over the state-of-the-art models.
\end{abstract}

\section{Introduction}
The advent of social media and its prosperity enable the creation of massive online user-generated content including opinions and product reviews. Analyzing such user-generated contents allows to detect the users' emotional states, which are useful for various downstream applications.

In the literature, there are a large number of works on emotion detection~\cite{RobersLREC2012,Abdul2017EmoNet,GuptaCoRR2017},
 both discrete and neural models have been used to predict the emotions of posts in social media. 
For example, \newcite{RobersLREC2012} used a series of binary SVM classifiers to detect the emotion of a post,  while \newcite{GuptaCoRR2017} used sentiment based semantic embedding and a LSTM model to learn the representation of a post for emotion detection.

Different from previous researches, which consider each post  individually, we think that posts in social media are much correlated by the authors' backgrounds.
Motivated by the \emph{principle of homophily}~\cite{lazarsfeld1954friendship}, the idea that similarity and connection tend to co-occur, or ``birds of a feather flock together'', suggests that users,  connected by mutual personal backgrounds, may hold similar opinions toward a post~\cite{thelwall2010emotion}.
In the literature, the personal attributes, such as gender, location, age, have been proved useful in personal background construction~\cite{Li2016Sentiment}: people with different attributes tend to express their emotions through different ways. For example, the \emph{happiness} emotion in [E1] is expressed through some femininity sense words, such as ``little brother'', ``handsome'', while the \emph{happiness} emotion in [E2] is expressed using a dialectal word ``bashi(\emph{comfortable})", which contains strong characteristic of the Sichuan dialect\footnote{Sichuan is a southwest province in China, Sichuan dialect(Sichuanese) is quite specific and different from standard Mandarin.}.  
Therefore, it is necessary to jointly detect the emotion of posts with the personal attributes.

\begin{quote}
	\textbf{[E1]} 祝贺中国男篮夺冠，新疆小哥哥，真是帅。 \\
	(\emph{Congratulations to Chinese basketball team on becoming the champion, the little brother of Xinjiang is really handsome.})\\
	
	\textbf{[E2]} 这种天气吃这个真是巴适！ \\
	(\emph{It is just bashi to eat this in such weather!})\\
\end{quote}
\vspace{-0.15 in}

However, the personal attributes are not easy to obtain in most social media websites. On one hand, most websites may not contain useful personal information.
On the other hand, people are normally not willing to attach their personal information in social media. 
Besides, integrating personal attributes into emotion detection is challenging, since it is hard to capture attributes-aware words, such as ``little brother'' and ``bashi''(\emph{comfortable}),  to connect the posts with similar backgrounds.
Although there are some related works on either personal attribute extraction~\cite{WangLSZ14} or emotion detection with personal attributes~\cite{Li2016Sentiment}, none of them address both challenges at the same time.
 
In this paper, we propose a Neural Personal Discrimination (NPD) model with both adversarial discriminators and attention mechanisms to tackle above challenges. 
Here, the \emph{Adversarial discriminators}~\cite{GoodfellowPMXWOCB14} are used to determine the personal attributes, e.g., gender or location, of a post, providing the inherent correlationship between emotions and personal backgrounds,
while the \emph{Attention mechanisms}~\cite{WangZLLZ16} are utilized to aggregate the representation of informative attributes-aware words into a vector for emotion prediction, providing insights into which words contribute to a personal background.
Experimental results show the usefulness of personal attributes in emotion detection, and the effectiveness of our proposed NPD model with both adversarial discriminators  and attention mechanisms over the state-of-the-art discrete and neural models.

\section{Related Work}
Earlier works on emotion detection are based on discrete models. For example, \newcite{yang2007} built a support vector machine (SVM) model and a conditional random field (CRF)  model for the emotion detection. \newcite{bhowmick2009multi} used a multi-label $k$NN model to classify a new sentence into multiple emotion categories. \newcite{DBLP:journals/tois/QuanWZSL15} proposed a logistic regression model for social emotion detection.
Recently, with the development of artificial intelligence, neural network models have been successfully applied to various NLP tasks~\cite{collobert2011natural,Goldberg2015A}. 
However,  few works use neural network models for emotion detection. \newcite{Abdul2017EmoNet} used a gated recurrent neural network model for emotion detection with a large-scale dataset. \newcite{zhang2018cross} used an auxiliary and attention based LSTM to detect emotion on a cross-lingual dataset.

%\subsection{Emotion Detection with Extra Resource}
Lexicon and social information are very important for emotion detection, and there are many researches focus on this topic.
For example, \newcite{strapparava2008learning} used WordNet-Affect to compute the sentimental score of a post.
More recently,
In addition, \newcite{hovy2015demographic} used both the age and gender information of the authors to improve the performance of sentiment analysis. 
\newcite{Vosoughi2015Enhanced} explored the relationship among locations, date time, authors and sentiments. 

Different from previous works which consider each post individually, we think that the posts in social media can be connected through the authors' backgrounds and should be better addressed.
On the basis, we propose a neural personal discrimination model to determine the personal background attributes from each post through adversarial discriminators, and aggregate the representation of informative attributes-aware words through attention mechanisms.
\section{Vanilla Model for Emotion Detection}

In this section, we propose a vanilla model. In the next section, we show how to utilize the neural personal discrimination model to improve the vanilla model by capturing personal attributes.
%which only considers text information for emotion detection, and treats emotion detection as a multi-label classification task~\cite{WangLLZ17}. 

In the vanilla model, we use a Long Short-Term Memory model to learn the representation of a post. On the basis, a multi-layer perception is learn to detect its emotions.

\subsection{Document Representation} \label{section-baseline}

In general, we denote a post as a document $d$ with $n$ words $\{w_1 ,w_2 ,..., w_n\}$.
Given the post, we use a standard Long Short-Term Memory (LSTM) model to learn the shared document representation. Specially, we transform each token $w_i$ into a real-valued vector $x_i$ using the word embedding vector of $w_i$, obtained by looking up a pre-trained word embedding table $D$ via the skip-gram algorithm to train embeddings~\cite{MikolovSCCD13}.
We then employ the LSTM model over $d$ to generate a hidden vector sequence $\{h_1,h_2,...,h_n\}$. At each step $t$, the hidden vector $h_t$ of the LSTM model is computed based on the current vector $x_t$ and the previous vector $h_{t-1}$ with $h_t=\LSTM(x_t,h_{t-1})$. The initial state and all stand LSTM parameters are randomly initialized and tuned during training. 

\subsection{Multi-label Emotion Detection}
Emotion detection aims to predict the emotion labels of posts. 
We follow \cite{WangZLLZ16} which adopts five kinds of emotions\footnote{The five emotions include \emph{happiness}, \emph{sadness}, \emph{anger}, \emph{fear}, and \emph{surprise}, please refer to ``Experiments Section'' of the paper for more details.} in the study.
Since there may be more than one emotion in a post, emotion detection can be considered as a multi-label classification task: we use $K$ emotion-specific binary perceptions ($K=5$) to predict if the post has the corresponding emotion or not.
%For each emotion, a binary classifier is trained using a multi-layer perception to predict the emotion label of the given post.
The advantage of multi-label classification is that it learns and predicts all the emotion labels jointly.

Formally, giving an input vector $H$, a hidden layer is first used to induce a set of high-level features for each emotion $j$:
\begin{equation}
\tilde{H}^j=\sigma (W^{j}H+b^{j}),
\end{equation}
and then, $H^j$ is used as inputs to a softmax output layer:
\begin{equation}
\hat{y}^j=\softmax(\tilde{W}^j\tilde{H}^j+\tilde{b}^j)
\end{equation}
Here, $W^{j}$, $b^{j}$, $\tilde{W}^j$, and $\tilde{b}^j$ are model parameters. 

\subsection{Training}
Given the word sequence in a post, our training objective is to minimize the cross-entropy loss over a set of training examples $(x_i, y_i)|_{i=1}^N$, with a $\ell_2$-regularization term,
\begin{equation}
\begin{split}
J_y(\theta_y)=-\sum_{i=1}^{N}\sum_{j=1}^Ky_i^j\log\hat{y}_i^j+\frac{\lambda}{2}||\theta_y||^2
\end{split}
\end{equation}
where $y_i^j$ represents the label of the $j$-th emotion for $x_i$,  $\theta_y$ is the set of model parameters and $\lambda$ is the parameter for $\ell_2$ regularization.

In this paper, the model parameters are optimized by AdaGrad~\cite{Duchi2011Adaptive}, and Skip-gram algorithm~\cite{MikolovSCCD13} is used for word embedding.

\section{Neural Personal Discrimination Model}

The drawback of above vanilla model is that it does not consider the deep personal correlationship among different posts.

In this study, we think that the posts in social media can be connected through the authors' backgrounds.
Therefore, we propose a Neural Personal Discrimination (NPD) model to connect people and learn their emotions collectively. 
%Since personal attributes are not easy to obtain in most social media websites, and it is hard to capture attributes-aware words from posts,
We use \emph{adversarial discriminators} to determine the personal attributes to construct the personal profiles, and employ \emph{attention mechanisms} to aggregate attributes-aware words. In this way, the social correlationship between different posts can be well addressed.

\begin{figure*}[!tp]
	\centering
	\includegraphics[width=380px]{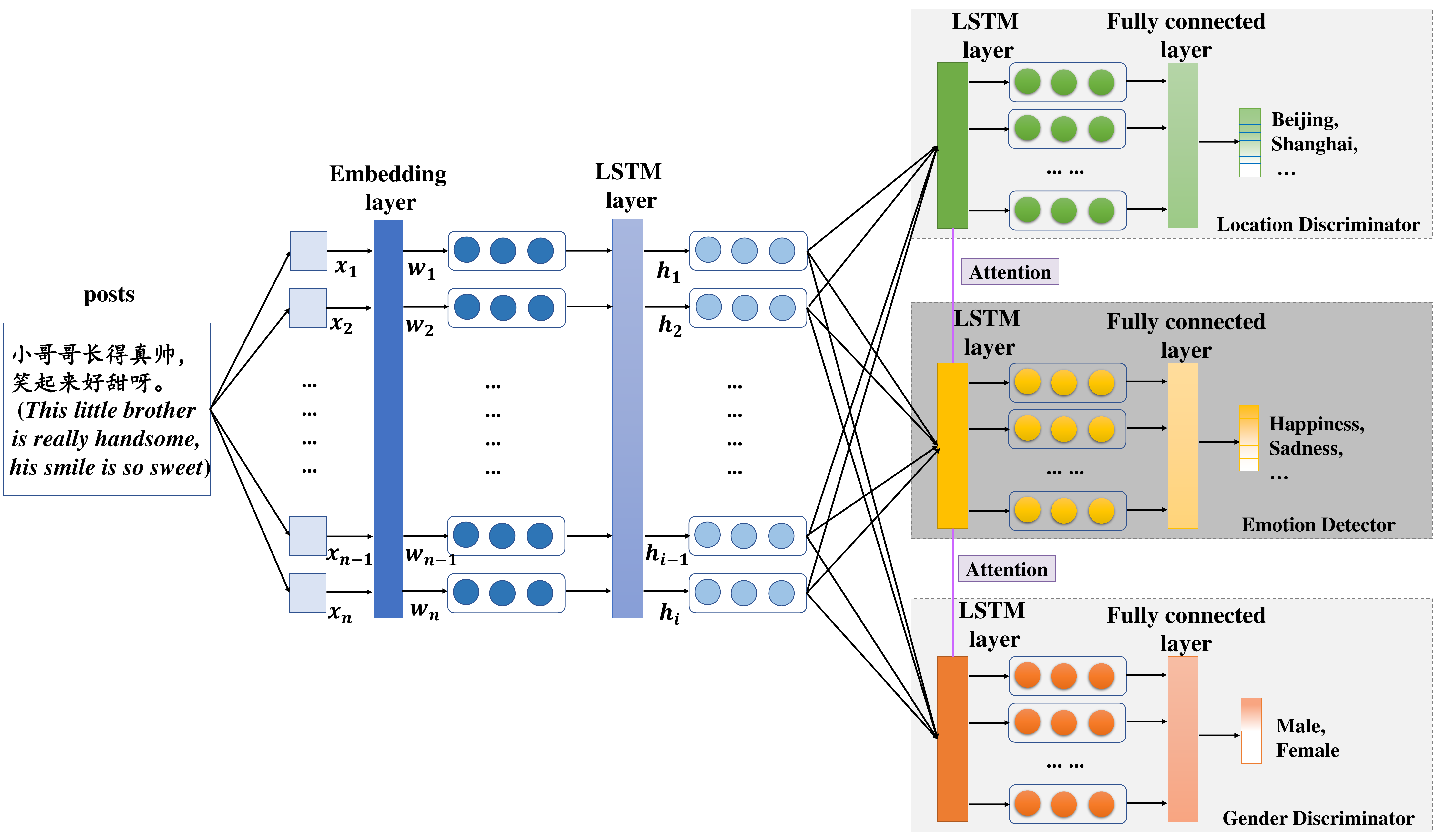}
	\caption{\label{figure-model} Overview of the neural personal discrimination model.}
\end{figure*}

Figure~\ref{figure-model} illustrates our proposed neural personal discrimination model for emotion detection. 
In particular, we first learn the representation of each post using a LSTM model, same as the vanilla joint model.
Then, we use adversarial discriminators to determine the personal attributes of each post. Finally, we employ attention mechanisms to aggregate the representation of informative attributes-aware words into a vector for the emotion prediction. 
In the following of this section, we illustrate the details of the infrastructure one by one.

\subsection{Personal Adversarial Discriminators}

A straightforward way to jointly detect personal attributes and the emotion of a post is to treat emotion detection as a multi-label classification.
However, such model may not be able to separate the posts from different attributes directly, and thus fail to learn the correlationship between the emotion of a post and the personal backgrounds of the authors. To address this issue, we utilize adversarial discriminators to determine the personal attributes of the authors, and to learn the emotion and the attributes of the authors collectively. 
Adversarial networks have achieved much success in various studies, especially in image and text generation~\cite{GoodfellowPMXWOCB14,Wang018,MaskGAN18}. 
In this part, we propose two adversarial discriminators, i.e, a gender discriminator and a location discriminator, to determine the personal attributes of each post.

\textbf{Gender Discriminator.}
The gender discriminator is employed to determine the author's gender of each post.
Let $g_i \in [0, 1]$ represents the probability of the gender label (female or male) for the gender discriminator, and $f$ is the function parameterized by $\theta_f$ which maps an embedding vector to a hidden representation $h_i^g$ from the post $x_i$.  
Here, the gender discriminator $G(h_i^g; \theta_g) \to \hat{g}_i$ parameterized by $\theta_g$ maps a hidden representation vector $h_i^g$ to a predicted gender label $\hat{g}_i$ with the loss function is:
\begin{equation}
J_{gend} = \sum_{i=1}^{N}\{g_i\log\hat{g}_i+(1-g_i)\log(1-\hat{g}_i)\},
\end{equation}
where $\hat{g}_i=G(f(x_i))$.

In this study, the gender discriminator is trained towards a saddle point of the loss function through maximizing the loss over $\theta_g$ while minimizing the loss over $\theta_f$~\cite{Ganin2015Domain}.

\textbf{Location Discriminator.}
The location discriminator is employed to determine the authors' location of a post\footnote{In this study, location discriminator is used to detect the author's province of a post.}. 
Let $\ell_i^j \in [0, 1]$ represents probability of the $j$-th location information of the $i$-th post and $j\in\{1,2,\dots, m\}$,  where, $m$ is the number of provinces.  The loss function is:
\begin{equation}
J_{loc} = \sum_{i=1}^N\sum_{j=1}^m\ell_i^j\log\hat{\ell}^j_i,
\end{equation}
where, $\hat{\ell}^j_i=L(f(x_i))$ and $L(h_i^{\ell}; \theta_{\ell})$ are the parameters of the location discriminator, $h_i^{\ell}=f(x_i)$ is the hidden represent from the post $x_i$

From the optimization of both discriminators, we can find that both $h^g$ and $h^{\ell}$ represent the latent feature representation of posts, which integrate the discrimination of various personal information. With the goal at $G(\theta_g)$ and $L(\theta_{\ell})$ try best to determine the gender and location of the authors. The adversarial network makes use of min-max optimization.   

\subsection{Personal Attention Mechanisms}
In  emotion detection, not all words contribute equally to the representation of emotions and personal attributes. 
Hence, we employ attention mechanisms to extract the words that are important to the personal backgrounds of posts, and to aggregate the representations of those informative attributes-aware words. 
With regard to the two adversarial discriminators, we propose attention mechanisms to build two representation ($v^g$ and  $v^{\ell}$) from the gender and location discriminators respectively, and then concatenate them together to construct the overall personal representation through the informative attributes-aware words .

\textbf{Gender Attention.}
We use an attention function to aggregate the gender-aware representation of the salient words to formulate the gender attention vector $v^g$.
Here, the gender attention model outputs a continuous vector $v^g \in \mathbb{R}^{d \times 1}$ recurrently by feeding the hidden representation vectors $\{h^g_1, h^g_2, \cdots, h^g_{n_t}\}$ as inputs. Specifically, $v^g$ is computed as a weighted sum of $h^g_i$ ($0\leq i\leq n_t$), namely
\begin{equation} \label{eq-9}
v^g=\sum_{i}^{n_t}\alpha_{i}h^g_i
\end{equation}
where $n_t$ is the hidden variable size, $\alpha_i \in [0,1]$ is the weight of $h^g_i$, and $\sum_i\alpha_i=1$. For each piece of hidden state $h^g_i$, the scoring function is calculated as follows:
\begin{equation} \label{eq-7}
\begin{split}
v_{i}^g=\tanh(W_gh^g_i+b^g)
\end{split}
\end{equation}
\begin{equation} \label{eq-beta}
\alpha_{i}=\frac{\exp(v^g)}{\sum_{j}{\exp(v_j^g)}}
\end{equation}

\textbf{Location Attention.}
Similar with the gender attention mechanism, the location attention model outputs a continuous vector $v^{\ell} \in \mathbb{R}^{d \times 1}$ recurrently by feeding the hidden representation vectors $\{h^{\ell}_1, h^{\ell}_2, \cdots, h^{\ell}_{n_k}\}$ as inputs. Specifically, $v^{\ell}$ is computed as a weighted sum of $h^{\ell}_i$ ($0\leq i\leq n_t$), namely
\begin{equation}
v^{\ell} = \sum_i^{n_k}\beta_{i}h^{\ell}_i 
\end{equation}
\begin{equation}
v^{\ell}_i = \tanh(W_{\ell}h^{\ell}_i + b^{\ell}) 
\end{equation}
\begin{equation}
\beta_i = \frac{\exp(v^\ell)}{\sum_{j}{\exp(v_j^{\ell})}} 
\end{equation}
where, $n_k$ is the location hidden variable size, and $\beta_i$ is the same setting as $\alpha_i$.

Finally, we concatenate $v^g$ and $v^{\ell}$ to capture the overall personal representation through all the personal attributes discriminators.
\begin{equation}
V = v^g \oplus v^{\ell}
\end{equation} 

\subsection{Adversarial Training with Neural Personal Discrimination}
The proposed NPD model can be trained in a end-to-end manner once we obtain the loss function of the emotion detector and the attribute discriminators. 
%We show that all the training processes can be embedded, Figure~\ref{figure-training} illustrates the process. The architecture includes a feature extractor (blue) and an emotion detector (red), which together form a standard deep LSTM architecture. Personal attributes are achieved by adding a gender discriminator (yellow) and a location discriminator (green), which are connected to the feature extractor via a gradient reversal layer that multiplies the gradient by a certain negative constant during the back propagation based training.
% \begin{figure}[!tp]
%	\centering
%	\includegraphics[width=210px]{training}
%	\caption{\label{figure-training} Adversarial training of neural personal discrimination model.}
%\end{figure}
Our ultimate training goal is to minimize the loss function with parameters $\theta=\{\theta_f, \theta_y, \theta_g, \theta_{\ell}\}$ as follow:
\begin{equation}\label{ANN loss}
J(\theta) = \lambda_1J_y + \lambda_2J_{gend} + \lambda_3J_{loc}
\end{equation}
Where$\lambda_1$, $\lambda_2$ and $\lambda_3$ are the weight parameters to balance the importance of losses between the emotion detection and the two personal attribute discriminators. 

Specifically, Eq.~\ref{ANN loss} is defined by finding a saddle point $\hat{\theta_y}, \hat{\theta}_f,  \hat{\theta}_g, \hat{\theta}_{\ell}$ such that
\begin{equation}\label{theta_f}
(\hat{\theta}_f,  \hat{\theta}_y)= \arg\min_{\theta_f, \theta_y} J(\theta_f,  \theta_y, \hat{\theta}_g, \hat{\theta}_{\ell})
\end{equation}
\begin{equation}\label{theta_g}
\hat{\theta}_g = \arg\max_{\theta_g }J(\hat{\theta}_f, \hat{\theta}_y, \theta_g, \hat{\theta}_{\ell})
\end{equation}
\begin{equation}\label{theta_l}
\hat{\theta}_{\ell} = \arg\max_{\theta_{\ell}} J(\hat{\theta}_f,  \hat{\theta}_y, \hat{\theta}_g, \theta_{\ell})
\end{equation}

As suggested previously, a saddle point is defined by Eq.~\ref{theta_f}$-$Eq.~\ref{theta_l}, and can be achieved as a stationary point the gradient updates:
\begin{equation}
\theta_f \gets \theta_f - \mu(\lambda_1\frac{\partial J_y^i}{\partial\theta_f}-\lambda_2\frac{\partial J_{gend}^i}{\partial\theta_f}-\lambda_3\frac{\partial J_{loc}^i}{\partial\theta_f})
\end{equation}
\begin{equation}
\theta_y \gets \theta_y - \mu\lambda_1\frac{\partial J_y^i}{\partial\theta_y}
\end{equation}
\begin{equation}
\theta_g \gets \theta_g- \mu\lambda_2\frac{\partial J_{gend}^i}{\partial\theta_g}
\end{equation}
\begin{equation}
\theta_{\ell} \gets \theta_{\ell}- \mu\lambda_3\frac{\partial J_{loc}^i}{\partial\theta_{\ell}}
\end{equation}
where $\mu$ is the learning rate.

\section{Experimentation}

\subsection{Experimental Settings} \label{section-data}

We collect the data from \emph{Weibo.com}, one of the most popular SNS websites in China. We crawl all the posts and corresponding personal profiles from the website. 
The dataset contains 11,157 microblog posts from 839 users. We employed six graduated students to annotate the corpus with a well-defined annotation guideline. Every two annotators annotate a same part of corpus, if they have disagreement on some posts, we ask another annotator to vote with them.
The annotation guideline is based on \newcite{Lee2013DETECTING}. 
Five basic emotions are annotated, namely \emph{happiness}, \emph{sadness}, \emph{fear}, \emph{anger} and \emph{surprise}~\cite{Lee2013DETECTING,WangLLZ17}. Table~\ref{tab:data} illustrates the statistics of each emotion. From the table, we find that the frequency of \emph{happiness} and \emph{sadness} are similar. Moreover, the frequency of \emph{fear} and \emph{anger} is much less than other three emotions.

%It may be due to that people always like to express their positive emotion in the social media.

\begin{table}[!tp]
	\caption{The statistics of emotion distribution in the dataset} \label{tab:data} \centering
	\vspace{-0.05 in}
	\begin{tabular}{c|c}
		\hline \bf Emtion & \bf Post Number \\
		\hline
		Happiness&2,915\\
		\hline
		Sadness&2,454\\
		\hline
		Fear&359\\
		\hline
		Anger&153\\
		\hline
		Surprise&601\\
		\hline
		None&4,675\\
		\hline
	\end{tabular}
\end{table}

We randomly select 70\% posts as the training data, and remaining 30\% posts as the test data. For evaluation, F1-measure is used to evaluate the performance of proposed model in each emotion. Average F1-measure is used to evaluate the overall performance of all emotions.  

The setting of hyper-parameters is: vocabulary size is 2000, batch size is set to 32, dropout ratio is 0.2, learning rate $\mu$ is set 0.0001, and $\lambda_1:\lambda_2:\lambda_3=1:1:1$

\begin{table*}[!tp]
	\caption{Experimental results of different models.} \label{tab:baseline} \centering
	\vspace{-0.05 in}
	\begin{tabular}{c|c|c|c|c|c|c}
		\hline
		\multirow{2}{*}{ \bf Method} & \multicolumn{6}{|c}{\bf F1.}\\
		\cline{2-7}
		&Happiness&Sadness&Anger&Surprise&Fear&Average\\
		\hline
		SVM	& 0.628	& 0.462 & 0.390& 0.117	& 0.091	 & 0.338\\
		\hline
		Abdul17 & 0.656 & 0.492 & 0.429& 0.103 & 0.111  & 0.358\\
		\hline
		Vaswani17 & 0.644 & 0.494 & 0.392 & 0.113 & 0.121 & 0.353\\ 
		\hline
		NPD & \bf 0.657 & \bf 0.510 & \bf 0.459 & \bf 0.135 & \bf 0.127  & \bf 0.378 \\
		\hline
	\end{tabular}
\end{table*}

\subsection{Experimental Results}
We compare the proposed Neural Personal Discrimination (NPD) model with several representative baselines models in Table~\ref{tab:baseline}, where,1) \textbf{SVM} is a widely used baseline to predict the emotion of a post in social media~\cite{yang2007}.2) \textbf{Abdul17} is a standard LSTM model which consist of a LSTM layer and a fully connected layer, and it is modified from the model in \newcite{Abdul2017EmoNet}. The LSTM model yields the state-of-the-art performance on emotion detection in recent researches.3)\textbf{Vaswani17} is an improved LSTM model with a self-attention mechanism. The self-attention mechanism is used to capture the structural information and has been successfully applied in various natural language processing tasks recently~\cite{cheng2016long,vaswani2017attention}

From Table~\ref{tab:baseline}, we find that all of the neural models outperform SVM significantly. This indicates that neural models are much more effective than discrete models in emotion detection.
In addition, our proposed NPD model outperforms both the standard LSTM model (Abdul17) and the improved LSTM model with self-attention (Vaswani17) significantly. 
This shows the effectiveness of our proposed NPD model with both adversarial discriminators and attention mechanisms. This also shows the usefulness of personal attributes for emotion detection.
Moreover, we find that the performance of Vaswani17 is even lower than the standard LSTM model. This shows that simply integrating a self-attention mechanism may not be able to well capture informative words for emotion detection in social media.

\subsection{Analysis and Discussion}
In this subsection, we analyze the influence of different factors in the proposed NPD model, and give some statistics and examples to illustrate the effectiveness of the proposed NPD model with different personal attributes.

\subsubsection{Influence of Personal Attributes}
We illustrate the influence of personal attributes in the proposed NPD model in Table~\ref{tab:factors}, where, 
1)\textbf{LSTM-attributes} is a LSTM based multi-label classification model, which predicts both the emotion and the attribute labels of each post collectively.2)\textbf{NPD-gender} ablates the location attribute, i.e., only considering the gender attributes in the NPD model.3) \textbf{NPD-location} ablates the gender attribute, i.e., only considering the location attributes in the NPD model.

From the table, we can find that the performance of  LSTM-attributes is much lower than  LSTM model. This indicates that simple multi-label classification setting is not effective for integrating personal attributes. This may be due to the fact that  basic multi-label setting fails to learn the correlationship between the emotions of posts and the personal attributes of the authors well. 
In addition, both the NPD-gender and the NPD-location perform better than the LSTM model respectively. This shows the effectiveness of the proposed NPD model with both adversarial and attention networks, and the usefulness of both gender and location attributes. 
Finally, the proposed NPD model with all the attributes significantly outperforms all the other models. This suggests that we should integrate all the personal attributes for emotion detection in social media.

\begin{table}[!tp]
	\caption{Comparison of various models with different personal attributes.} \label{tab:factors} \centering
	\vspace{-0.05 in}
	\begin{tabular}{c|c}
		\hline
	    \bf{Method}&\bf{Average F1.}\\
		\hline
		LSTM&0.358\\
		\hline
		LSTM-attributes&0.341\\ 
		\hline
		NPD-gender&0.360\\
		\hline
		NPD-location&0.363\\
		\hline
		NPD&\bf 0.378\\
		\hline
	\end{tabular}
\end{table}
 
\subsubsection{Influence of Network Structures}
After analyzing the influence of different attributes, we analyze the influence of network structures in Table~\ref{tab:configuration}, where \textbf{LSTM-attention} ablates the adversarial discriminators, and only utilizes attention mechanisms with a multi-label classification setting, and \textbf{LSTM-adversarial} ablates the attention mechanisms and only utilizes adversarial discriminators for emotion detection.

From Table~\ref{tab:configuration}, we can see that both the attention mechanisms (LSTM-attention) and the adversarial discriminators (LSTM-adversarial) are effective in emotion detection. Moreover, the adversarial discriminators are much more effective than the attention mechanisms. This shows that the personal attribute discriminator is more important than learning informative attributes-aware words. Moreover, the NPD model obtains the best results by integrating both adversarial discriminators and attention mechanisms.

\begin{table}[!tp]
	\caption{Influence of network structures.} \label{tab:configuration} \centering
	\vspace{-0.05 in}
	\begin{tabular}{c|c}
		\hline
	    \bf{Method}&\bf{Average F1.}\\
		\hline
		LSTM&0.358\\
		\hline
	    LSTM-attention&0.363\\
		\hline
		LSTM-adversarial&0.375\\
		\hline
	    NPD&\bf{0.378}\\
		\hline
	\end{tabular}
\end{table}

\subsubsection{Statistics}

We give the statistics of gender and location to explore the correlationship between emotion and personal attributes.

\begin{figure}[!tp]
	\centering
	\includegraphics[width=220px]{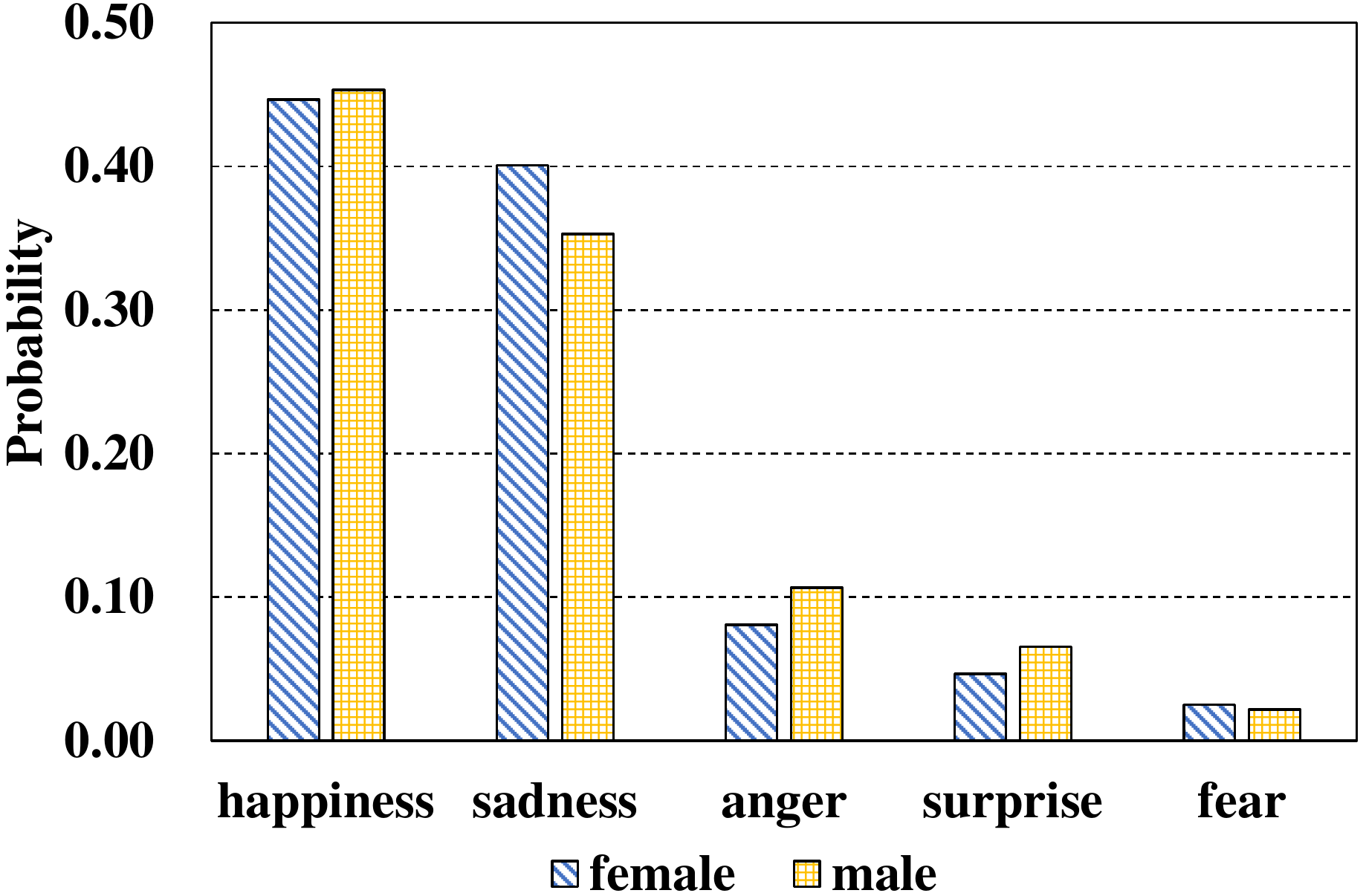}
	\caption{\label{figure-gender} Distribution of gender.}
\end{figure}

\textbf{Distribution of Gender.} Figure~\ref{figure-gender} illustrates the distribution of emotions between genders. Here, the Y-axis is conditional probability of each emotion given gender. 
From the figure, we can find that women tend to express the sadness emotion, while men tend to express anger emotion. This may be due to the fact that the different personality has different emotion expressions, i.e., sentimentality of the female and the impulsion of the male.

\begin{figure}[t]
	\centering
	\includegraphics[width=220px]{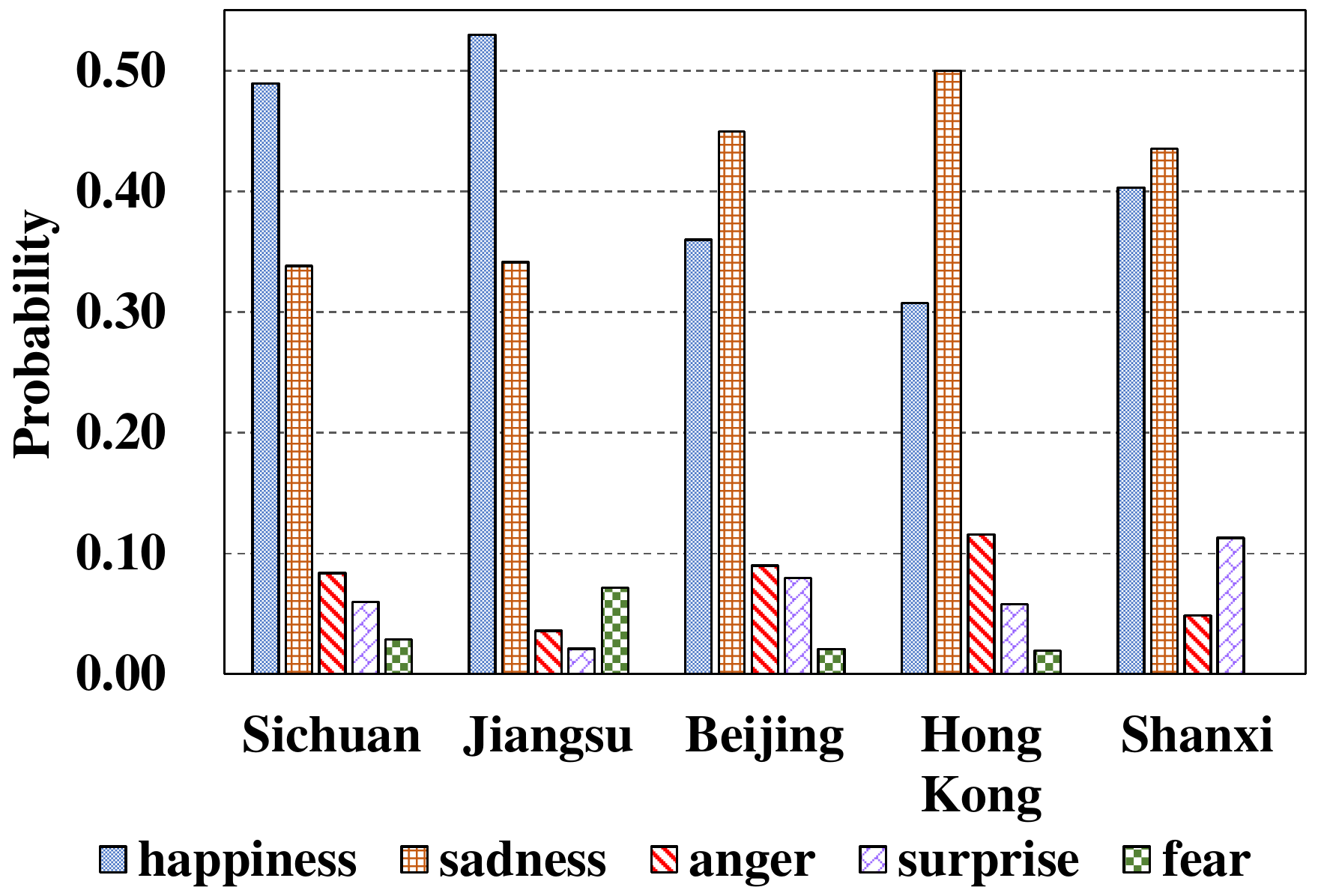}
	\caption{\label{figure-location} Distribution of location.}
\end{figure}
\begin{table*}[t]
	\newcommand{\tabincell}[2]{\begin{tabular}{@{}#1@{}}#2\end{tabular}}
	\caption{Examples predicted labels of LSTM and NPD} \label{tab:case} \centering
	\vspace{-0.05 in}
	\begin{tabular}{p{11cm}|c|c}
		\hline
		\bf{Post}\centering&\bf{LSTM}&\bf{NPD}\\
		\hline
		\tabincell{l}{[E3]@嬤嬤俊, 连九份芋圆都会做, 巴适惨了, 冰冰凉Q弹水果甜品.\\(\emph{@Momo Jun, this guy can even make Taro Balls from Jiufen of }\\ \emph{Taiwan. It is so bash. This ice fruit dessert is so soft and chewy.})}&Sadness&Happiness\\
		\hline
		\tabincell{l}{[E4]下期节目, 不容错过, 有我的男神出场.\\(\emph{Next tv show is not to be missed, because my Mr. Mcdreamy}\\ \emph{ will be coming.})}&None&Happiness\\
		\hline
		\tabincell{l}{[E5]我这身体不敢恭维啊, 吃了好多的冰, 结果痛经了吧, 赶紧多喝\\点红糖姜茶.\\(\emph{I can’t compliment my body. Eating too much ice leads to my}\\ \emph{  dysmenorrhea. I must to drink more brown sugar ginger tea})}&None&Sadness\\
		\hline
	\end{tabular}
\end{table*}
\textbf{Distribution of Location.}  Figure~\ref{figure-location} is an example of the distribution of emotions between locations.
As discussed in the above section, we use the province of authors as location attributes.
Here, the Y-axis represents the conditional probability of each emotion given location. From the figure, we can find that the authors' location can may influence their emotions in many aspects. For example,
people tend to express the positive (i.e., happiness) emotion than the negative emotion in Jiangsu. One of the most comfortable and developed regions in China.
Due to air pollution and populations, people tend to express the negative emotion than the positive emotion in Beijing.
In addition, people in Hong Kong always feel crowding and tend to express the sadness emotion.
Finally, people always feel happy and comfortable in Sichuan, well known as ``country of paradise'' in China.

From the results of statistics, we can see that the personal attributes, like gender, location, can affect the emotion detection. Also, the experimental results of personal attribute influence shows the same conclusion. 

\subsubsection{Case Study}
We select three examples from the test set to evaluate the effectiveness of the proposed NPD model for better comparison with the LSTM model in Table~\ref{tab:case}.
In [E3], through ``bashi(\emph{conformable})'' is a strong cue word of the \emph{happiness} emotion, it is treated as a general word without indicating any location information from the LSTM model. At the same time, the location information (Sichuan Province) determined by the location discriminator and the attention mechanism makes ``bashi(\emph{conformable})'' critical in the NPD model, and these enable the NPD model determines the correct \emph{happiness} emotion. 
In [E4] and [E5], both posts contain implicit gender information. For example, ``Mr. Mcdreamy'' implies the female's affection for the handsome male, and ``dysmenorrhea" is a female physiological disease. Without the background of such gender information, it is impossible to infer any potential emotional information from these two words. This explains why the LSTM model fails to detect any emotion from the examples (\emph{None} of emotion). 
However, our NPD model successfully determines the gender of these two posts by the gender discriminator, and improves the weight of these two words for emotion detection by the attention mechanisms. In this way, the emotions of the three examples are correctly detected through the proposed NPD model.

\section{Conclusion}
Most of previous studies consider each post individually in emotion detection, one of the most important tasks in sentiment analysis. However, 
since the posts in social media are generated by users, it is natural that these posts can be connected through authors' personal background attributes.
In this paper, we propose a neural personal discrimination model with both adversarial discriminators and attention mechanisms to connect posts with similar personal attributes. 
Here, the adversarial discriminators are used to determine the personal attributes, and the attention mechanisms are employed to aggregate attributes-aware words. In this way the social correlations between different posts can be captured.
The experimental results show the usefulness of the personal attributes and the effectiveness of our proposed neural personal discrimination model  in modeling such personal attributes with significant performance improvement over the state-of-the-art baselines.

\section*{Acknowledgment}
This work was support by National Natural Science Foundation of China (No. 61702518, No.61806137).
This work was also supported by the joint research project of Alibaba and Soochow University.
Finally, we would like to thank the anonymous reviewers for their insightful comments and suggestions.
\bibliography{emnlp19_1}

\begin{thebibliography}{29}
\expandafter\ifx\csname natexlab\endcsname\relax\def\natexlab#1{#1}\fi

\bibitem[{Abdul{-}Mageed and Ungar(2017)}]{Abdul2017EmoNet}
Muhammad Abdul{-}Mageed and Lyle~H. Ungar. 2017.
\newblock \href {https://doi.org/10.18653/v1/P17-1067} {Emonet: Fine-grained
  emotion detection with gated recurrent neural networks}.
\newblock In \emph{ACL}, pages 718--728.

\bibitem[{Bhowmick et~al.(2009)Bhowmick, Basu, Mitra, and
  Prasad}]{bhowmick2009multi}
Plaban~Kumar Bhowmick, Anupam Basu, Pabitra Mitra, and Abhishek Prasad. 2009.
\newblock \href {https://doi.org/10.1007/978-3-642-11164-8\_42} {Multi-label
  text classification approach for sentence level news emotion analysis}.
\newblock In \emph{ICPRAI}, pages 261--266.

\bibitem[{Cheng et~al.(2016)Cheng, Dong, and Lapata}]{cheng2016long}
Jianpeng Cheng, Li~Dong, and Mirella Lapata. 2016.
\newblock \href {http://aclweb.org/anthology/D/D16/D16-1053.pdf} {Long
  short-term memory-networks for machine reading}.
\newblock pages 551--561.

\bibitem[{Collobert et~al.(2011)Collobert, Weston, Bottou, Karlen, Kavukcuoglu,
  and Kuksa}]{collobert2011natural}
Ronan Collobert, Jason Weston, L{\'{e}}on Bottou, Michael Karlen, Koray
  Kavukcuoglu, and Pavel~P. Kuksa. 2011.
\newblock \href {http://dl.acm.org/citation.cfm?id=2078186} {Natural language
  processing (almost) from scratch}.
\newblock \emph{Journal of Machine Learning Research}, 12:2493--2537.

\bibitem[{Colneri{\^c} and Demsar(2018)}]{colneric2018emotion}
Niko Colneri{\^c} and Janez Demsar. 2018.
\newblock Emotion recognition on twitter: Comparative study and training a
  unison model.
\newblock \emph{IEEE Transactions on Affective Computing}, PP(99):1--14.

\bibitem[{Duchi et~al.(2011)Duchi, Hazan, and Singer}]{Duchi2011Adaptive}
John~C. Duchi, Elad Hazan, and Yoram Singer. 2011.
\newblock \href {http://dl.acm.org/citation.cfm?id=2021068} {Adaptive
  subgradient methods for online learning and stochastic optimization}.
\newblock \emph{Journal of Machine Learning Research}, 12:2121--2159.

\bibitem[{Fedus et~al.(2018)Fedus, Goodfellow, and Dai}]{MaskGAN18}
William Fedus, Ian~J. Goodfellow, and Andrew~M. Dai. 2018.
\newblock Maskgan: Better text generation via filling in the
  {\_}{\_}{\_}{\_}{\_}{\_}{\_}.
\newblock In \emph{ICLR}, pages 1--16.

\bibitem[{Ganin et~al.(2017)Ganin, Ustinova, Ajakan, Germain, Larochelle,
  Laviolette, Marchand, and Lempitsky}]{Ganin2015Domain}
Yaroslav Ganin, Evgeniya Ustinova, Hana Ajakan, Pascal Germain, Hugo
  Larochelle, Fran{\c{c}}ois Laviolette, Mario Marchand, and Victor~S.
  Lempitsky. 2017.
\newblock \href {https://doi.org/10.1007/978-3-319-58347-1\_10}
  {Domain-adversarial training of neural networks}.
\newblock pages 189--209.

\bibitem[{Goldberg(2016)}]{Goldberg2015A}
Yoav Goldberg. 2016.
\newblock \href {https://doi.org/10.1613/jair.4992} {A primer on neural network
  models for natural language processing}.
\newblock \emph{J. Artif. Intell. Res.}, 57:345--420.

\bibitem[{Goodfellow et~al.(2014)Goodfellow, Pouget{-}Abadie, Mirza, Xu,
  Warde{-}Farley, Ozair, Courville, and Bengio}]{GoodfellowPMXWOCB14}
Ian~J. Goodfellow, Jean Pouget{-}Abadie, Mehdi Mirza, Bing Xu, David
  Warde{-}Farley, Sherjil Ozair, Aaron~C. Courville, and Yoshua Bengio. 2014.
\newblock \href {http://papers.nips.cc/paper/5423-generative-adversarial-nets}
  {Generative adversarial nets}.
\newblock In \emph{NIPS}, pages 2672--2680.

\bibitem[{Gupta et~al.(2017)Gupta, Chatterjee, Srikanth, and
  Agrawal}]{GuptaCoRR2017}
Umang Gupta, Ankush Chatterjee, Radhakrishnan Srikanth, and Puneet Agrawal.
  2017.
\newblock \href {http://arxiv.org/abs/1707.06996} {A
  sentiment-and-semantics-based approach for emotion detection in textual
  conversations}.
\newblock \emph{CoRR}, abs/1707.06996.
\newblock Version 4.

\bibitem[{Hovy(2015)}]{hovy2015demographic}
Dirk Hovy. 2015.
\newblock \href {http://aclweb.org/anthology/P/P15/P15-1073.pdf} {Demographic
  factors improve classification performance}.
\newblock In \emph{ACL}, pages 752--762.

\bibitem[{Lazarsfeld et~al.(1954)Lazarsfeld, Merton
  et~al.}]{lazarsfeld1954friendship}
Paul~F Lazarsfeld, Robert~K Merton, et~al. 1954.
\newblock Friendship as a social process: A substantive and methodological
  analysis.
\newblock \emph{Freedom and control in modern society}, 18(1):18--66.

\bibitem[{Lee et~al.(2013)Lee, Chen, Huang, and Li}]{Lee2013DETECTING}
Sophia Yat~Mei Lee, Ying Chen, Chu{-}Ren Huang, and Shoushan Li. 2013.
\newblock \href {https://doi.org/10.1111/j.1467-8640.2012.00459.x} {Detecting
  emotion causes with a linguistic rule-based approach}.
\newblock \emph{Computational Intelligence}, 29(3):390--416.

\bibitem[{Li et~al.(2016)Li, Yang, and Zong}]{Li2016Sentiment}
Junjie Li, Haitong Yang, and Chengqing Zong. 2016.
\newblock \href {https://doi.org/10.1007/978-3-319-50496-4\_52} {Sentiment
  classification of social media text considering user attributes}.
\newblock In \emph{NLPCC/ICCPOL}, pages 583--594.

\bibitem[{Mikolov et~al.(2013)Mikolov, Sutskever, Chen, Corrado, and
  Dean}]{MikolovSCCD13}
Tomas Mikolov, Ilya Sutskever, Kai Chen, Gregory~S. Corrado, and Jeffrey Dean.
  2013.
\newblock \href
  {http://papers.nips.cc/paper/5021-distributed-representations-of-words-and-phrases-and-their-compositionality}
  {Distributed representations of words and phrases and their
  compositionality}.
\newblock In \emph{NIPS}, pages 3111--3119.

\bibitem[{Quan et~al.(2015)Quan, Wang, Zhang, Si, and
  Liu}]{DBLP:journals/tois/QuanWZSL15}
Xiaojun Quan, Qifan Wang, Ying Zhang, Luo Si, and Wenyin Liu. 2015.
\newblock \href {https://doi.org/10.1145/2749459} {Latent discriminative models
  for social emotion detection with emotional dependency}.
\newblock \emph{{ACM} Trans. Inf. Syst.}, 34(1):2:1--2:19.

\bibitem[{Roberts et~al.(2012)Roberts, Roach, Johnson, Guthrie, and
  Harabagiu}]{RobersLREC2012}
Kirk Roberts, Michael~A. Roach, Joseph Johnson, Josh Guthrie, and Sanda~M.
  Harabagiu. 2012.
\newblock \href
  {http://www.lrec-conf.org/proceedings/lrec2012/summaries/201.html}
  {Empatweet: Annotating and detecting emotions on twitter}.
\newblock In \emph{LREC}, pages 3806--3813.

\bibitem[{Strapparava and Mihalcea(2008)}]{strapparava2008learning}
Carlo Strapparava and Rada Mihalcea. 2008.
\newblock \href {https://doi.org/10.1145/1363686.1364052} {Learning to identify
  emotions in text}.
\newblock In \emph{SAC}, pages 1556--1560.

\bibitem[{Thelwall(2010)}]{thelwall2010emotion}
Mike Thelwall. 2010.
\newblock \href
  {http://firstmonday.org/htbin/cgiwrap/bin/ojs/index.php/fm/article/view/2897}
  {Emotion homophily in social network site messages}.
\newblock \emph{First Monday}, 15(4):1--8.

\bibitem[{Vaswani et~al.(2017)Vaswani, Shazeer, Parmar, Uszkoreit, Jones,
  Gomez, Kaiser, and Polosukhin}]{vaswani2017attention}
Ashish Vaswani, Noam Shazeer, Niki Parmar, Jakob Uszkoreit, Llion Jones,
  Aidan~N. Gomez, Lukasz Kaiser, and Illia Polosukhin. 2017.
\newblock \href {http://papers.nips.cc/paper/7181-attention-is-all-you-need}
  {Attention is all you need}.
\newblock In \emph{NIPS}, pages 6000--6010.

\bibitem[{Vosoughi et~al.(2016)Vosoughi, Zhou, and Roy}]{Vosoughi2015Enhanced}
Soroush Vosoughi, Helen Zhou, and Deb Roy. 2016.
\newblock \href {http://arxiv.org/abs/1605.05195} {Enhanced twitter sentiment
  classification using contextual information}.
\newblock \emph{CoRR}, abs/1605.05195.
\newblock Version 1.

\bibitem[{Wang and Wan(2018)}]{Wang018}
Ke~Wang and Xiaojun Wan. 2018.
\newblock \href {https://doi.org/10.24963/ijcai.2018/618} {Sentigan: Generating
  sentimental texts via mixture adversarial networks}.
\newblock In \emph{IJCAI}, pages 4446--4452.

\bibitem[{Wang and Zheng(2013)}]{wang2013text}
Xuren Wang and Qiuhui Zheng. 2013.
\newblock Text emotion classification research based on improved latent
  semantic analysis algorithm.
\newblock In \emph{ICCSEE}, pages 210--213.

\bibitem[{Wang et~al.(2017)Wang, Lee, Li, and Zhou}]{WangLLZ17}
Zhongqing Wang, Sophia Yat~Mei Lee, Shoushan Li, and Guodong Zhou. 2017.
\newblock \href {https://doi.org/10.1109/TASLP.2016.2637280} {Emotion analysis
  in code-switching text with joint factor graph model}.
\newblock \emph{{IEEE/ACM} Trans. Audio, Speech {\&} Language Processing},
  25(3):469--480.

\bibitem[{Wang et~al.(2014)Wang, Li, Shi, and Zhou}]{WangLSZ14}
Zhongqing Wang, Shoushan Li, Hanxiao Shi, and Guodong Zhou. 2014.
\newblock \href {http://aclweb.org/anthology/C/C14/C14-1050.pdf} {Skill
  inference with personal and skill connections}.
\newblock In \emph{COLING}, pages 520--529.

\bibitem[{Wang et~al.(2016)Wang, Zhang, Lee, Li, and Zhou}]{WangZLLZ16}
Zhongqing Wang, Yue Zhang, Sophia Yat~Mei Lee, Shoushan Li, and Guodong Zhou.
  2016.
\newblock \href {http://aclweb.org/anthology/C/C16/C16-1153.pdf} {A bilingual
  attention network for code-switched emotion prediction}.
\newblock In \emph{COLING}, pages 1624--1634.

\bibitem[{Yang et~al.(2007)Yang, Lin, and Chen}]{yang2007}
Changhua Yang, Kevin~Hsin{-}Yih Lin, and Hsin{-}Hsi Chen. 2007.
\newblock \href {https://doi.org/10.1109/WI.2007.51} {Emotion classification
  using web blog corpora}.
\newblock In \emph{WI-IAT}, pages 275--278.

\bibitem[{Zhang et~al.(2018)Zhang, Wu, Li, Wang, and Zhou}]{zhang2018cross}
Lu~Zhang, Liangqing Wu, Shoushan Li, Zhongqing Wang, and Guodong Zhou. 2018.
\newblock \href {https://doi.org/10.1007/978-3-319-99495-6\_36} {Cross-lingual
  emotion classification with auxiliary and attention neural networks}.
\newblock In \emph{NLPCC}, pages 429--441.

\end{thebibliography}
\bibliographystyle{acl_natbib}

\end{CJK}
\end{document}